\documentclass{article} % For LaTeX2e
\usepackage{iclr2015,times}
\usepackage{amssymb}
\usepackage{hyperref}
\usepackage{url}
\usepackage{graphicx}
\usepackage{multirow} 

\title{Deep Networks with Large Output Spaces}

\author{
Sudheendra Vijayanarasimhan, Jonathon Shlens, Rajat Monga \& Jay Yagnik \\
Google Research\\
Mountain View, CA 94043, USA \\
\texttt{\{svnaras,shlens,rajatmonga,jyagnik\}@google.com} \\
}

\iclrfinalcopy

\begin{document}

\maketitle

\begin{abstract}
Deep neural networks have been extremely successful at various image, speech, video recognition tasks because of their ability to model
deep structures within the data. However, they are still prohibitively expensive to train and apply for problems containing millions of
classes in the output layer. Based on the observation that the key computation common to most neural network layers is a vector/matrix product,
we propose a fast locality-sensitive hashing technique to approximate the actual dot product enabling us to scale up the training and inference
to millions of output classes. We evaluate our technique on three diverse large-scale recognition tasks and show that our approach can train 
large-scale models at a faster rate (in terms of steps/total time) compared to baseline methods.
\end{abstract}

\section{Introduction}
\label{sec:intro}
Deep neural networks have proven highly successful at various image, speech, language and video recognition
 tasks~\citep{krizhevsky2012imagenet,mikolov13,karpathy2014large}. These networks typically have several layers
of units connected in feedforward fashion between the input and output spaces. Each layer performs a specific function
 such as convolution, pooling, normalization, or plain matrix products in the case of fully connected layers followed by
 some form of non-linear activation such as sigmoid or rectified linear units.

Despite their attractive qualities, and the relative efficiency of their local architecture, these networks are still 
prohibitively expensive to train and apply for large-scale problems containing millions of classes or nodes. 
There are several such problems proposed in the literature. The Imagenet dataset which is one of the largest datasets for image classification contains around
 $21000$ classes. Wordnet, which is a superset of Imagenet, consists of $117,00$ synsets. Freebase, which is a
 community-curated database of well-known people, places, and things contains close to $20$ million entities. Image models of text queries have ranged from
 $100000$ queries in academic benchmarks~\citet{wsabie} to several million in commercial search engines such as Google, Bing, Yahoo, etc. Duplicate video content
identification~\citep{shang, song, zhao} and video recommendations are also large-scale problems with millions of classes.

We note that the key computation common to softmax/logistic regression layers is a matrix product between the activations from a layer, $x$, and 
the weights of the connections to the next layer, $W$. As the number of classes increases this computation becomes the main bottleneck of the entire network. 
Based on this observation, we exploit a fast locality-sensitive hashing
 technique~\citep{YagniketalICCV-11} in order to approximate the actual dot product in the final output layer which enables us to
 scale up the training and inference to millions of output classes. 
 
%NOTE (SHLENS): WE NEED to DIFFERENTIATE HERE   .. AS A FIRST STEP, WE CAN EMPLOY THIS TECHNIQUE WHEN
%WHEN A NETWORK IS DOMINATED BY THE CLASSIFICATION LAYER. THIS HAPPENS WHEN THE CARDINALITY IS QUITE LARGE. HERE WE EXPLORE THIS
%TOPICS IN THE PAPER .. (I.E. WE MAKE NO ATTEMPT TO SPEED UP CONVOLUTIONS, ETC. OR INTERNAL WORKINGS OF A NETWORK).

Our main idea is to approximate the dot product between the output layer's parameter vector and the input activations using hashing. We first
 compute binary hash codes for the parameter vectors, $W$, of a layer's output nodes and store the indices of the nodes in locations
 corresponding to the hash codes within hash tables. During inference, given an input activation vector, $x$, we compute the hash codes
 of the vector and retrieve the set of output nodes $O_k$ that are closest to the input vector in the hash space. Following this we compute
 the actual dot product between $x$ and the parameter vectors of $O_k$ and set all other values to zero.

By avoiding the expensive dot product operation between the input activation vector and all output nodes we show that our approach can
easily scale up to millions of output classes during inference. Furthermore, using the same technique when training the models,
we show that our approach can train large-scale models at a faster rate both in terms of number of steps and the total time compared to both
the standard softmax layers and the more computationally efficient hierarchical softmax layer of~\citep{mikolov13}.

\section{Related Work}
\label{sec:related}
Several methods have been proposed for performing classification in deep networks
over large vocabularies. Traditional methods such as logistic regression and
softmax (multinomial regression) are known to have poor scaling properties
with the number of classes~\citep{dean13} as the number of dot products grows
with the number of classes C that must be considered.

One method for contending with this is hierarchical softmax whereby a tree is
constructed of depth $log_2 C$ in which the leaves are the individual classes
which must be classified~\citep{morin05, mikolov13}. A benefit of this
approach is that each  step merely requires computations associated with
the tree traversal to an individual leaf.

A second direction is to instead train a dense embedding space representation
and perform classification by employing k-nearest-neighbors in this embedding
space on unseen examples. Typical methods for training such embedding representations
employ a hinge rank loss with a clever selection of negative examples, e.g.~\citep{weston11}.

Locality sensitive hashing (LSH)~\citep{GionisetalVLDB-99} provides a third alternative by providing
methods to perform approximate nearest neighbor search in sub-linear time for various similarity
metrics. An LSH scheme based on ordinal similarity is proposed in~\citep{YagniketalICCV-11} which is
used in \citep{dean13} to speed-up filter based object detection. We expand on these techniques 
to enable learning large-scale deep network models.

\section{Approach}
\label{sec:approach}

The goal of this work is to enable approximate computation of the matrix product of the parameters of a layer and its input activations,
$x^T W$, in a deep network so that the number of output dimensions can be increased by several orders of magnitude. 
In the following sections, we demonstrate that a locality sensitive hashing based approximation can provide such a solution without too
 much degradation in overall accuracy. As a first step we employ this technique to scale up the final classification layer since the benefits of hashing
are easily seen when the cardinality is quite large.

\subsection{Softmax/Logistic Classification}

Softmax and logistic regression functions are two popular choices for the final layer of a deep network  
for multi-class and binary classification problems respectively. Formally, the two functions are defined as 

\begin{eqnarray}
  P_{softmax}(y = j | x) &=& \frac{e^{x^{T}w_j}}{\sum_{k = 1}^{N} e^{x^{T}w_k}} \\
  P_{logistic}(y = j | x) &=& \frac{1}{1 + e^{-(x^{T}w_j + \beta_{j})}} \\
\label{eq:softmax}
\end{eqnarray}

where $P(y = j | x)$ is the probability of the $j^{th}$ class given the input vector $x$ and $\{w_{j}, j = 1 ... N\}$ are distinct 
linear functions for the $N$ classes.

When the number of classes is large, not all classes are relevant to a given input example. Therefore, in many situations we are only interested
 in the $K$ classes with the highest probabilities. We could obtain the top $K$ classes by equivalently determining the $K$ vectors, $W_{K}$,
 that have the largest dot products with the input vector $x$ and computing the probabilities for only these $K$ classes, setting all others to zero. 

We note that this is equivalent to the problem of finding the approximate nearest neighbors of a vector based on cosine (dot product) similarity 
which has a rich literature beginning with the seminal work of~\citep{GionisetalVLDB-99}. It has been shown that approximate nearest neighbors can be obtained 
in time that is sub-linear in the number of database vectors with certain guarantees which is the key motivation for our approach.

In our case the database vectors are the parameter vectors of the output layer, $w_j$, and the query vector is the input activation
from the previous layer $x$. In this work, we employ the subfamily of hash functions, \emph{winner-take-all} (WTA) hashing introduced in~\citep{YagniketalICCV-11},
 since it has been successfully applied for the similar task of scaling up filter-based object detection in~\citep{dean13}.

\begin{figure}
\centering
\includegraphics[width=0.7\linewidth]{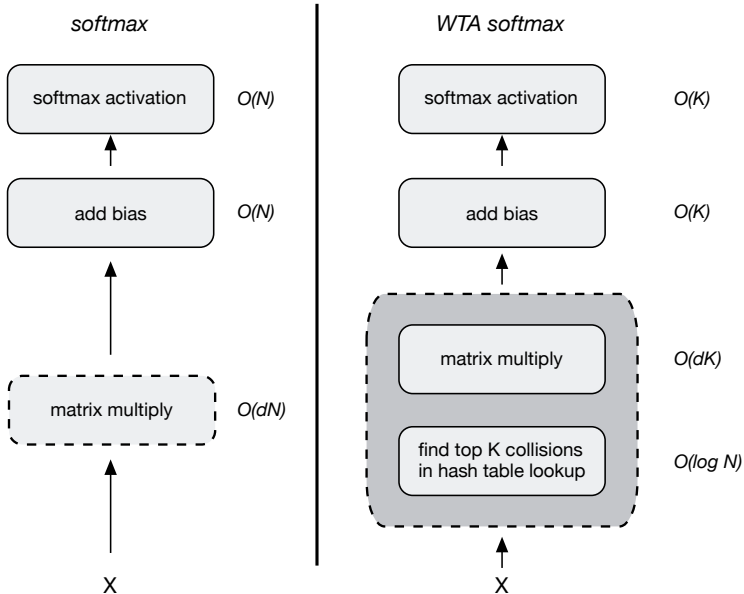}
\caption{\small A diagram comparing a typical classification network trained with softmax with the proposed WTA softmax. The left column shows the operations of softmax $ = f(WX + b)$. $X$ are the input network activations, $f(\cdot)$ is the softmax activation function, $b$ are biases for each of $N$ classes and $W$ is an $N\times d$ matrix where each row is the weight vector associated with an individual class. The matrix product $WX$ is the most expensive operation for the entire network when the number of classes $N$ is extremely large. The right column diagrams the WTA softmax operation. The hashing operation identifies the $K << N$ most likely labels for a given $X$. The remainder of the WTA softmax operations are largely identical although they only operate on the $K << N$ likely labels.}
%\includegraphics[width=0.8\linewidth]{figs/wta_final_layer.eps}
%\caption{A schematic describing the main idea behind our approach. The learned parameter vectors $W$ are stored in hash tables using WTA hashing and the hash codes
%of the input vector $x$ are used to retrieve the top $K$ classes with the largest dot products with the input vector. The actual dot product and the corresponding
%probabilites are computed for only these retrieved classes.}
\label{fig:wta}
\end{figure}

\subsection{Winner-Take-All Hashing (WTA)}
\label{sec:wta}
Given a vector $x$ or $w$ in $\mathbb{R}^d$, its WTA hash is defined by permuting its elements using $P$ distinct permutations and recording the
 index of the maximum value of the first $k$ elements~\citep{YagniketalICCV-11}. Each index can be compactly represented using $log_2 k$ bits resulting
in $P * log_2 k$ bits for the entire hash. The WTA hash has several desirable properties; since the only operation involved in
computing the hash is comparison, it can be completely implemented using integer arithmetic and the algorithm can be efficiently 
implemented without accruing branch prediction penalties.

Furthermore, each WTA hash function defines an ordinal embedding and it has been shown in~~\citep{YagniketalICCV-11} that as $P \to d!$,
 the dot product between two WTA hashes tends to the rank correlation between the underlying vectors. Therefore, WTA is well suited as a basis
for locality-sensitive hashing as ordinal similarity can be used as a more robust proxy for dot product similarity.

Given binary hash codes, $u$ of a vector, $w$, there are several schemes that can be employed in order to perform approximate nearest neighbor 
search. In this work, we employ the scheme used in~\citep{dean13} due to its simplicity and limited overhead.

In this scheme, we first divide the compact hash code, $u$, containing $P$ elements of $log_2 k$ bits into $M$ bands, $u_m$, each containing $P / M$ elements.
 We create a hash table for each band $\{T_m, m = 1 ... M\}$ and store the index of the vector in the hash bins corresponding to $u_m$ in each hash table $T_m$. During 
retrieval, we similarly compute the hash codes of $x$ and divide it into $M$ bands and retrieve the set of all IDs in the corresponding hash bins
along with their counts. The counts provide a lower bound for the dot product between the two hash vectors which is related to the ordinal similarity between the
two vectors. Therefore, the top $K$ IDs from this list approximate the $K$ nearest neighbors to the input vector based on dot product similarity. 
The actual dot product can now be computed for these vectors with the input vector $x$ to obtain their probabilities.

The complexity of the scheme proposed above depends on the dimensionality of the vectors for computing the hash codes, the number of bands or hash tables 
that are used during retrieval, $M$, and the number of IDs for which the actual dot product is computed, $K$. Since all three quantities are independent of the 
number of classes in the output layer our approach can accomodate any number of classes in the output layer. As shown in Figure~\ref{fig:wta}, the naive softmax
 has a complexity of $O(d * N + N)$ whereas our WTA based approximation has a complexity of $O(d * K + K + constant)$. The overall speed-up we obtain is of the
 order of $\frac{N}{K}$ assuming the cost of computing the hash function and lookup are much smaller than $d * N$. Of course, since both $M$ and $K$ relate to
 the accuracy of the approximation they provide a trade-off between the time complexity and the accuracy of the network.

\subsection{Inference}

We can apply our proposed approximation during both model inference and training. For inference, given a learned model, we first compute the hash codes of the 
parameter vectors of the softmax/logistic regression layer and store the IDs of the corresponding classes in the hash table as described in Section~\ref{sec:wta}.
This is a one time operation that is performed before running inference on any input examples.

Given an input example, we pass it through all layers leading up to the classification layer as before and compute the hash codes of the input activations
to the classification layer. We then query the hash table to retrieve the top $K$ classes and compute probabilties using Equation~\ref{eq:softmax} for only
 these classes. Figure~\ref{fig:wta} shows a rough schematic of this procedure.

\subsection{Training}

\begin{figure}
\centering
\includegraphics[width=0.7\linewidth]{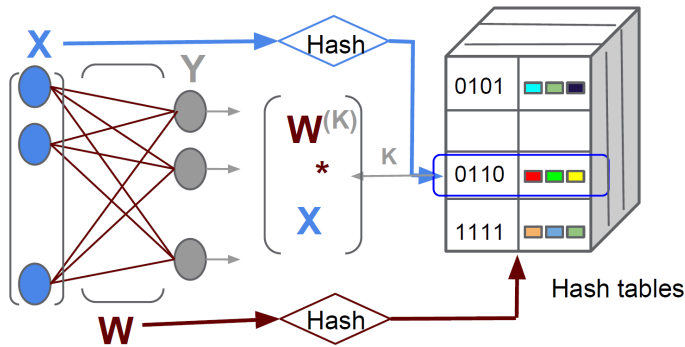}
\caption{\small A schematic describing the use of the hash table during inference and training. The learned parameter vectors $W$ are stored in hash tables using
 WTA hashing and the hash codes of the input vector $x$ are used to retrieve the top $K$ classes with the largest dot products with the input vector.
 The actual dot product and the corresponding probabilites are computed for only these retrieved classes. Similarly, during the backward pass the gradients
are computed based on the top $K$ retrieved nodes and the parameter vectors are updated.}
%\includegraphics[width=0.8\linewidth]{figs/update-scheme.eps}
%\caption{A schematic describing the interactions between the WTA layer and other components of the model. The WTA layer queries the hash table for the top $K$ most
%relevant classes for the input $x$ and obtains the parameter vectors for these and the $U$ classes that need to be updated in this round. It then computes the 
%probabilities for the retrieved nodes and additionally updates the hash tables using the new parameter vectors. Similarly, during the backward pass the gradients
%are computed based on the top $K$ retrieved nodes and the parameter vectors are updated.}
\label{fig:update-scheme}
\end{figure}

We train the models using downpour SGD, an asynchronous stocastic gradient descent procedure supporting a large number of model replicas, proposed in~\citep{dean12}. During backpropagation we only propagate gradients based on the top $K$ classes that were retrieved during the forward pass of the model and update the
 parameter vectors of only these retrieved classes using the error vector. Additionally, we add all the positive labels for an input example to the list of
 non-zero classes in order to always provide a positive gradient. In Section~\ref{sec:experiments} we show empirical results of performing only these top $K$
 updates. These sparse gradients are much more computationally efficient and additionally perform the function of hard negative mining since only the closest
 classes to a particular example are updated.

While inference using WTA hashing is straightforward to implement, there are several challenges that need to be solved to make training efficient using such a scheme.
Firstly, unlike during inference, the parameter vectors are constantly changing as new examples are seen. It would be infeasible to request updated
parameters for all classes and update the hash table after every step. 

However, we found that gradients based on a small set of examples do not perturb the parameter vector significantly and WTA hashing is only sensitive to changes in
 the ordering of the various dimensions and is more robust to small changes in the absolute values of the different dimensions. Based on these observations we 
implemented a scheme where the hash table locations of classes are updated in batches in a round-robin fashion such that all classes are updated over a course of
 several hundred or thousand steps which turned out to be quite effective.

Therefore, we only request updated parameters for the set of retrieved classes, the positive training classes and the classes selected in the round-robin scheme.
Figure~\ref{fig:update-scheme} shows a schematic of these interactions with the parameter server and the hash tables.

\section{Experiments}
\label{sec:experiments}
We empirically evaluate our proposed technique on several large-scale datasets with the aim of
investigating the trade-off between accuracy and time complexity of the WTA based softmax classifier
in comparison to the baseline approaches of softmax (exhaustive) and hierarchical softmax.

\subsection{Imagenet 21K}

The 2011 Imagenet 21K dataset consists of 21,900 classes and 14 million images. We split the set into
equal partitions of training and testing sets as done in~\citep{le12}. We selected values of $16$, $1000$, $3000$ for
the $k$, $M$, $P$ parameters of the WTA approach for all experiments based on results on a small set of images which
agreed with the parameters mentioned in~\citep{dean13}. We varied the value of $K$, which is the number of 
retrieved classes for which the actual dot product is computed, since it directly affects both the accuracy
and the time complexity of the approach.

We used the artichecture proposed in~\citep{krizhevsky2012imagenet} (AlexNet) for all experiments replacing only the 
classification layer with the proposed approach. All methods were optimized using downpour SGD with a starting learning
rate of 0.001 with exponential decay in conjunction with a momentum of 0.9. We used a cluster of about $100$ machines
containing multi-core CPUs with $20 GB$ of RAM running at $2.4 Ghz$ to perform training and inference.

\begin{figure}
\begin{minipage}{0.45\textwidth}
\scalebox{0.75}{
\begin{tabular}{|c|c|c|c|c|}
\hline
\multirow{2}{*}{Batch Size} & \multirow{2}{*}{K} & \multicolumn{2}{c|}{Time (ms)} & \multirow{2}{*}{Speedup}\\
& & WTA  & Softmax & \\
\hline
1 & 30 &        1.8 & \multirow{4}{*}{128.2} & 71.2x \\
1 & 300 &       3.0 & & 42.7x \\
%1 & 1000 &      4.6 & & 27.9x \\
1 & 3000 &      9.4 & & 13.6x \\
\hline
8 & 30 &        2.3 & \multirow{4}{*}{158.2} & 68.8x \\
8 & 300 &       4.2 & & 37.7x \\
%8 & 1000 &      9.0 & & 17.6x \\
8  & 3000 &     23.0 & & 6.9x  \\
\hline
32 & 30 &       6.7 & \multirow{4}{*}{210.7} & 31.3x \\
32 & 300 &     13.2 &  & 16.0x \\
%32 & 1000 &    29.5 &  & 7.1x \\
32 & 3000 &    77.9 & & 2.7x \\
\hline
64 & 30 &      13.3 & \multirow{4}{*}{277.2} & 20.8x \\
64 & 300 &     25.2 &  & 11x \\
%64 & 1000 &    58.4 &  & 4.7x \\
64 & 3000 &     173.1 &  & 1.6x \\
%\hline
%128 & 50 &     27.4 & \multirow{4}{*}{427.1} & 15.6x \\
%128 & 300 &    45.7 &  & 9.3x \\
%128 & 1000 &  115.6 &  & 3.7x \\
%128 & 3000   & 315.6 &  & 1.35x \\
\hline 
\end{tabular}}
\caption{\small Time taken by the WTA softmax layer and regular softmax layer alone for various values of batch size and top $K$ for a prediction space of 21K classes
 during inference. WTA provides significant speed-upds over softmax for small batch sizes and small values of $K$. Note that due to the sublinear nature of hash 
retreival, the speedups will be larger for bigger problems. }
\label{tab:time}
\end{minipage}\hspace{0.5in}
%% \begin{minipage}{0.5\textwidth}
%% \centering
%% \begin{tabular}{|c|c|c|c|c|}
%% \hline
%% K & \multicolumn{2}{c|}{Accuracy}\\
%% & WTA & Softmax \\
%% \hline
%% 30 & 23.2 & \multirow{3}{*}{27.9} \\ 
%% 300 & 25.2 & \\
%% 3000 & 27.3 & \\
%% \hline 
%% \end{tabular}
%% \caption{}
%% \label{tab:acc}
%% \end{minipage}
\begin{minipage}{0.45\textwidth}
\centering
\includegraphics[width=\linewidth]{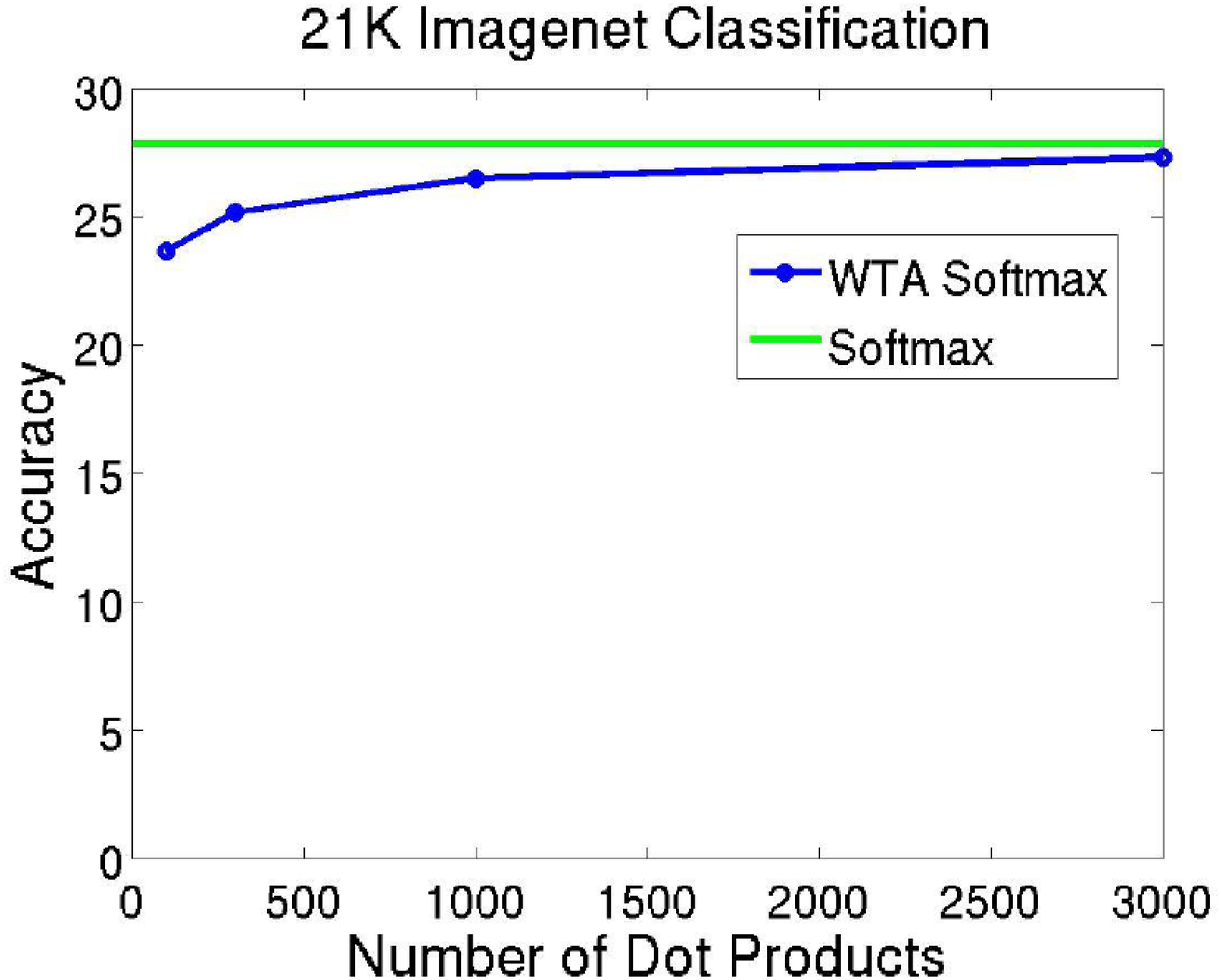}
\caption{\small Accuracies obtained by the WTA model as the number of retrieved classes, $K$, is varied from 30 to 3000. Even with as few as 30 classes the WTA
model is able to reach $83\%$ of the accuracy of the baseline model and almost reaches the baseline accuracy for $K = 3000$. Note that this result uses WTA to
 just approximate an already trained network and hence the ceiling is the accuracy of the base network.}
\label{fig:imagenet-vs-dot}
\end{minipage}
\end{figure}

Figure~\ref{tab:time} first reports the time taken during inference by the WTA Softmax and Softmax layers alone (ignoring the rest of the model) as
 both the batch size and the top $K$ is varied for this problem. We note that WTA Softmax provides significant speed-up over Softmax
for both small batch sizes and small values of $K$. For large batch sizes Softmax is very efficient due to optimizations
over dense matrices. For large values of $K$ the dot product with the retrieved vectors begins to dominate the time 
complexity.

Figure~\ref{fig:imagenet-vs-dot} report the accuracies obtained when using WTA during inference on a learned model as compared to
the baseline accuracy of the softmax model. We find that even with as few as $30$ retrieved classes our approach is able to
reach up to $83\%$ of the baseline accuracy and almost matches the baseline accuracy with $3000$ retrieved classes. Note that the ceiling on this problem 
is the accuracy of the base network since we are approximating an already trained network using WTA. This vindicates our 
claim that only a small percentage of classes are relevant to any input example and WTA hashing provides an efficient
technique for obtaining the top $K$ most relevant classes for a given input example. 
Based on these figures we conclude that the proposed approach is advantageous when either $N$ is very large or for small batch sizes. 

\begin{figure}
\begin{minipage}[t]{0.45\textwidth}
\centering
\includegraphics[width=0.8\linewidth]{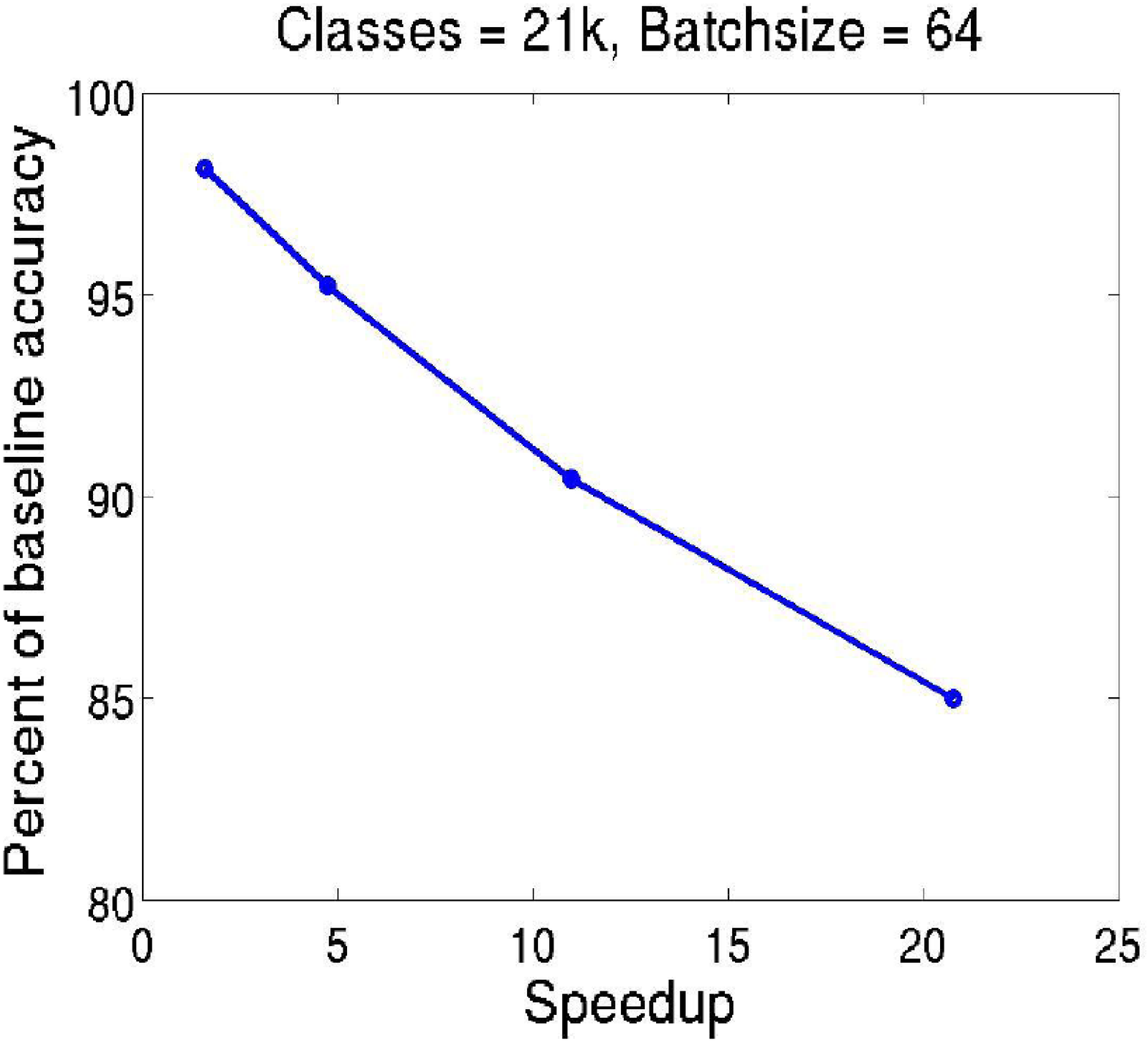}
\caption{\small Trade-off between the speed-up achieved over the baseline softmax model at a fixed batch size and the percentage of the baseline accuracy reached.
WTA achieves a speed-up of 10x at $90\%$ of the baseline accuracy. }
\label{fig:imagenet-speedup}
\end{minipage}\hspace{0.3in}
\begin{minipage}[t]{0.45\textwidth}
\centering
\includegraphics[width=0.8\linewidth]{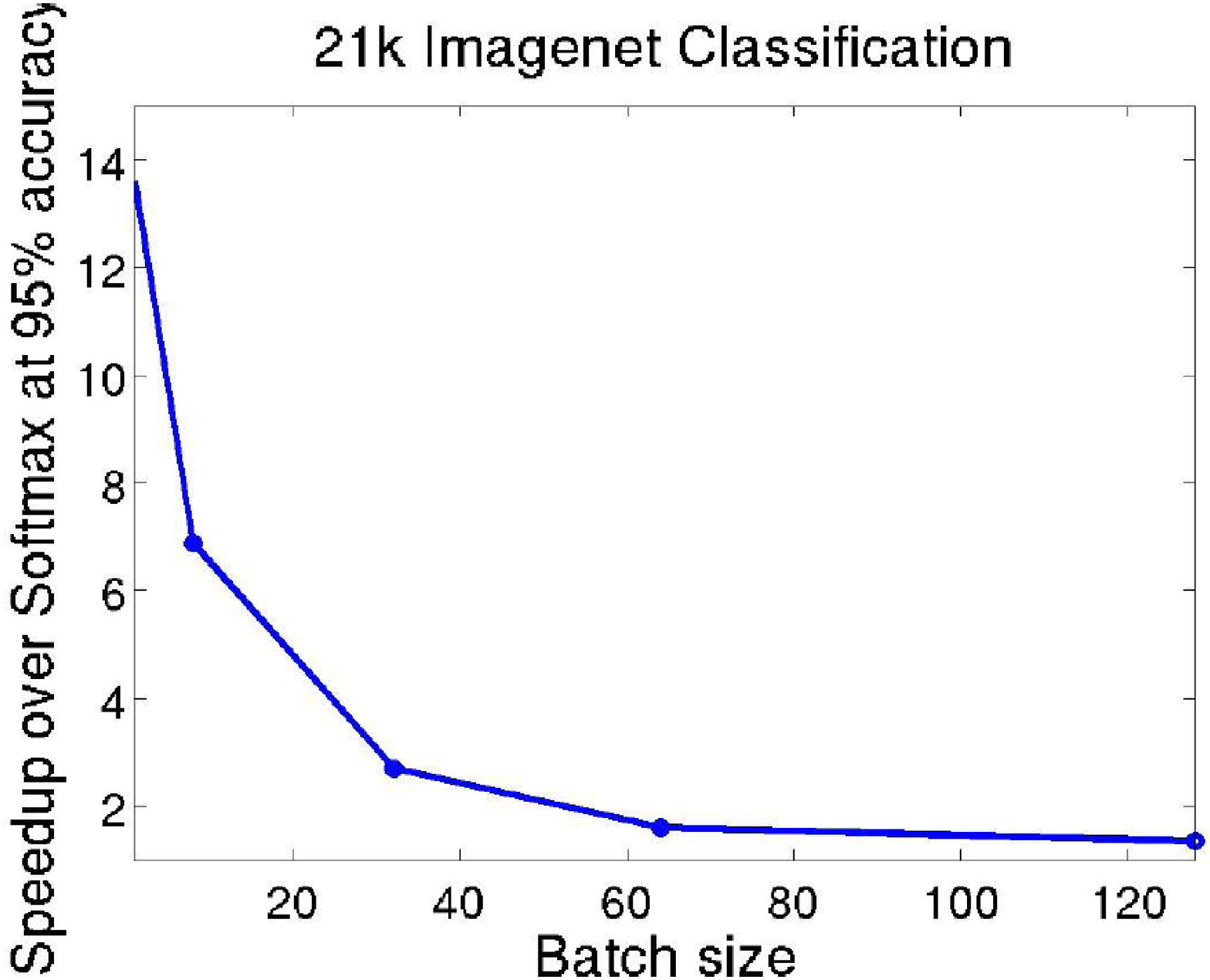}
\caption{\small Speed-up achieved by WTA at $95\%$ of the baseline accuracy at various batch sizes.}
\label{fig:imagenet-batchsize}
\end{minipage}
\end{figure}

Figure~\ref{fig:imagenet-speedup} reports the trade-off between the speedup achieved over baseline softmax at a fixed batch size and the percentage of the 
baseline accuracy reached by the WTA model. We find that the WTA model achieves a speedup of 10x over the baseline model at $90\%$ accuracy.
Figure~\ref{fig:imagenet-batchsize} reports the speedup achieved at $95\%$ of the baseline accuracy for various batch sizes. As noted previously we find that
the WTA model achieves higher speedups for smaller batch sizes.

\subsection{Skipgram dataset}

\label{sec:language-model-experiment}
One popular application of deep networks has been building and training models of
language and semantics. Recent work from ~\citep{mikolov13,quoc} has demonstrated that
a shallow, simple architecture can be trained efficiently by across language corpora.
The resulting embedding vector representation for language exhibits
rich structure about semantics and syntactics that can be exploited for many
other purposes. For all of these models, a crucial aspect of training is to predict
surrounding and nearby words in a sequence. The prediction task is typically quite large, i.e.
the cardinality is the size of the vocabulary, O(1M-10M) words.

A key insight of recent work has been to exploit novel and efficient methods for
performing discrete classification over large cardinalities. In particular, said work
employs a hierarchical softmax to fast inference and evaluation.

As a test of the predictive performance our hashing techniques, we compare
the performance of WTA hashing on the language modeling task. We note that this
is an extremely difficult task the perplexity of language (or just cooccurrence statistics
of words in a sentence) is quite high. Thus, any method attempts to predict nearby words
will at best report low predictive performances.

In our experiments, we download and parse Wikipedia consisting of several billion sentences.
We tokenize this text corpora with the 1M most popular words. The task of the network is
to perform a 1M-way prediction nearby words based on neighboring words.

We performed our experiments with three loss functions, traditional softmax, hierarchical softmax
and WTA-based softmax. We found measure the precision@K for the top K predictions from each softmax
model.

We compare all networks with three loss functions after 100 hours of training time across similar
CPU time. We find that all networks have converged within this time frame although the hierarhical
softmax has processed 100 billion examples while the WTA softmax has processed 100 million examples.

%TODO(shlens): Rerun results with traditional softmax. Rerun with cross validation.

As seen in Table~\ref{tab:skipgram}, we find that WTA softmax achieves superior predictive performance than the hierarchical softmax
even though hierarchical softmax has processed O(100) times more examples. In particular, we find that
WTA softmax achieves roughly two-fold better predictive performance.

%We note that although the WTA softmax model outperforms hierarchical softmax in predictive performance
%with far fewer training examples,
However, the WTA softmax produces underlying embedding vector representations
that do not perform as well on analogy tasks as highlighted by ~\citep{mikolov13}. For instance, the hierarchical
softmax achieves 50\% accuracy on analogy tasks where as WTA softmax produces 5\% accuracy on similar tasks.
This is partly due to the fewer number of examples processed by WTA in the same time frame as hierarchical softmax
is significantly faster than WTA because it performs just $log(N)$ dot products.

%TODO(shlens): Expand on analogy issue and backpropagation more.

%Thus, although WTA softmax does not work well for a training useful embedding vector representations (which is
%the main goal of the skipgram language model), these experiments do demonstrate that WTA softmax is notably better
%at discrete classification tasks in which the cardinality is extremely large, i.e. O(1 million). We take these
%results as proof of concept for efficient learning and prediction using a hashing based approximation on problems of
%large cardinality.

%TODO(shlens): Make this into a LaTeX table. Just text right now.

\begin{table}
\centering
\begin{tabular}{|c|c|c|}
\hline                        
     & H-Softmax &       WTA-Softmax \\
\hline
precision@1 &		1.15\%	&	1.93\% \\
precision@3 &		2.36\% &		5.18\% \\
precision@5 &		3.14\%		& 7.48\% \\
precision@10 &		4.52\%		&10.2\% \\
precision@20 &		6.28\%	&	13.4\% \\
precision@50 &		9.63\%		& 16.5\% \\
precision@100 &		13.2\% &		18.5\% \\
\hline
average precision &	2.31\%  &          4.53\% \\
\hline
\end{tabular}
\caption{The precison@k values of WTA softmax and hierarchical softmax models on the prediction problem of the skipgram dataset. WTA softmax achieves higher
predictive accuracy even though it processes much fewer training examples in the alloted time.}
\label{tab:skipgram}
\end{table}

\subsection{Video Identification}

While the 21K problem is one of the largest tested for the baseline softmax model, the benefits of hashing are best seen for problems of much larger cardinality.
In order to illustrate this we next consider a large-scale classification task for video identification. This task is modeled on Youtube's content ID
 classification problem which has also been addressed in several recent work under various settings~\citet{shang,song,zhao}.

The task we propose is to predict the ID of a video based on its frames.  We use the Sports 1M action recognition dataset introduced
 in~\citep{karpathy2014large} for this problem. The Sports 1M dataset consists of roughly 1.2 million Youtube sports videos annotated with 487 classes. 
We divide the first five minutes of each video into two parts where the first $50\%$ of the video's frames are used for training and the remaining $50\%$ are
 used for evaluating the models. The prediction space of the problem spans 1.2 million classes and each class has roughly 150 frames for training and
 evaluation. 

We trained three models for this problem with the AlexNet architecture where the top layer uses one of softmax, WTA softmax and hierarchical softmax each. We
used learning rates of $\{0.1, 0.05, 0.001, 0.005\}$ and report the best results for each of the models. For WTA we used a value of 3000 for the $K$ parameter
based on the results in the previous section and a batch size of $32$ for all models.

\begin{figure}
\begin{minipage}{0.3\textwidth}
\centering
\includegraphics[width=\linewidth]{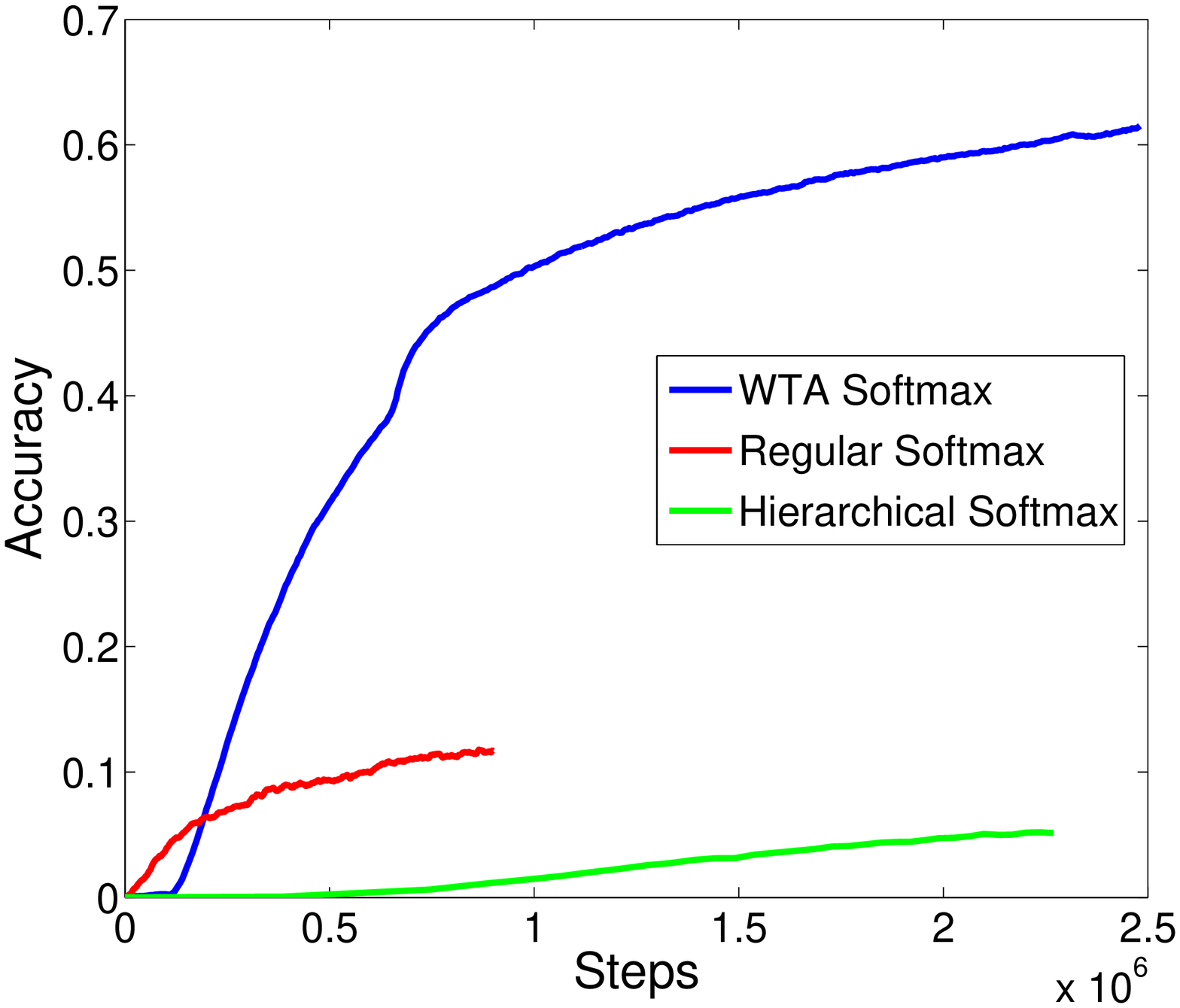}
\caption{\small Accuracy of the various models on the Sports 1M evaluation set as a function of the number of training steps.}
\label{fig:sports-vs-steps}
\end{minipage}\hspace{0.1in}
\begin{minipage}{0.3\textwidth}
\centering
\includegraphics[width=\linewidth]{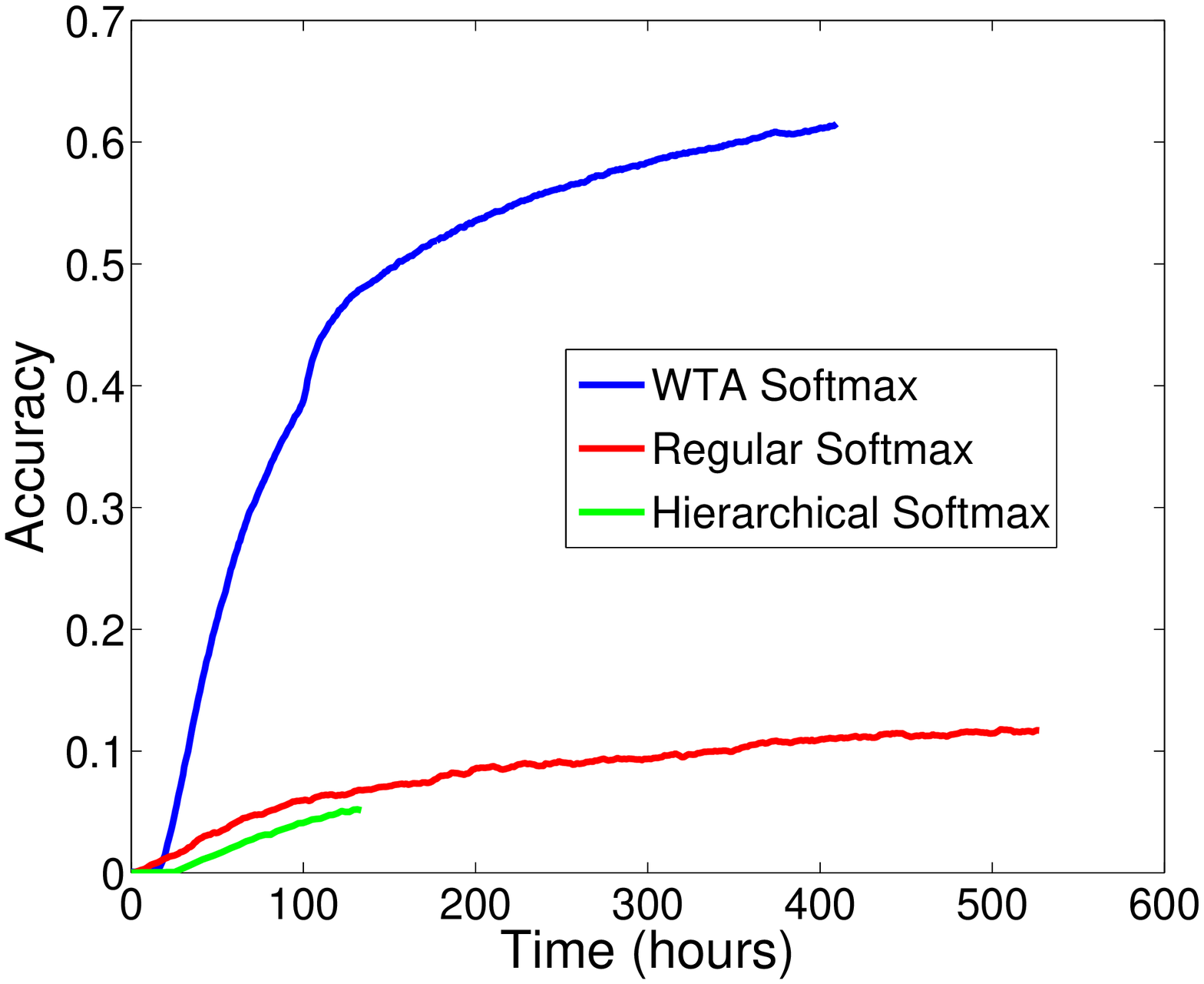}
\caption{\small Accuracy of the various models on the Sports 1M evaluation set as a function of the total training time. }
\label{fig:sports-vs-time}
\end{minipage}\hspace{0.1in}
\begin{minipage}{0.3\textwidth}
\centering
\includegraphics[width=\linewidth]{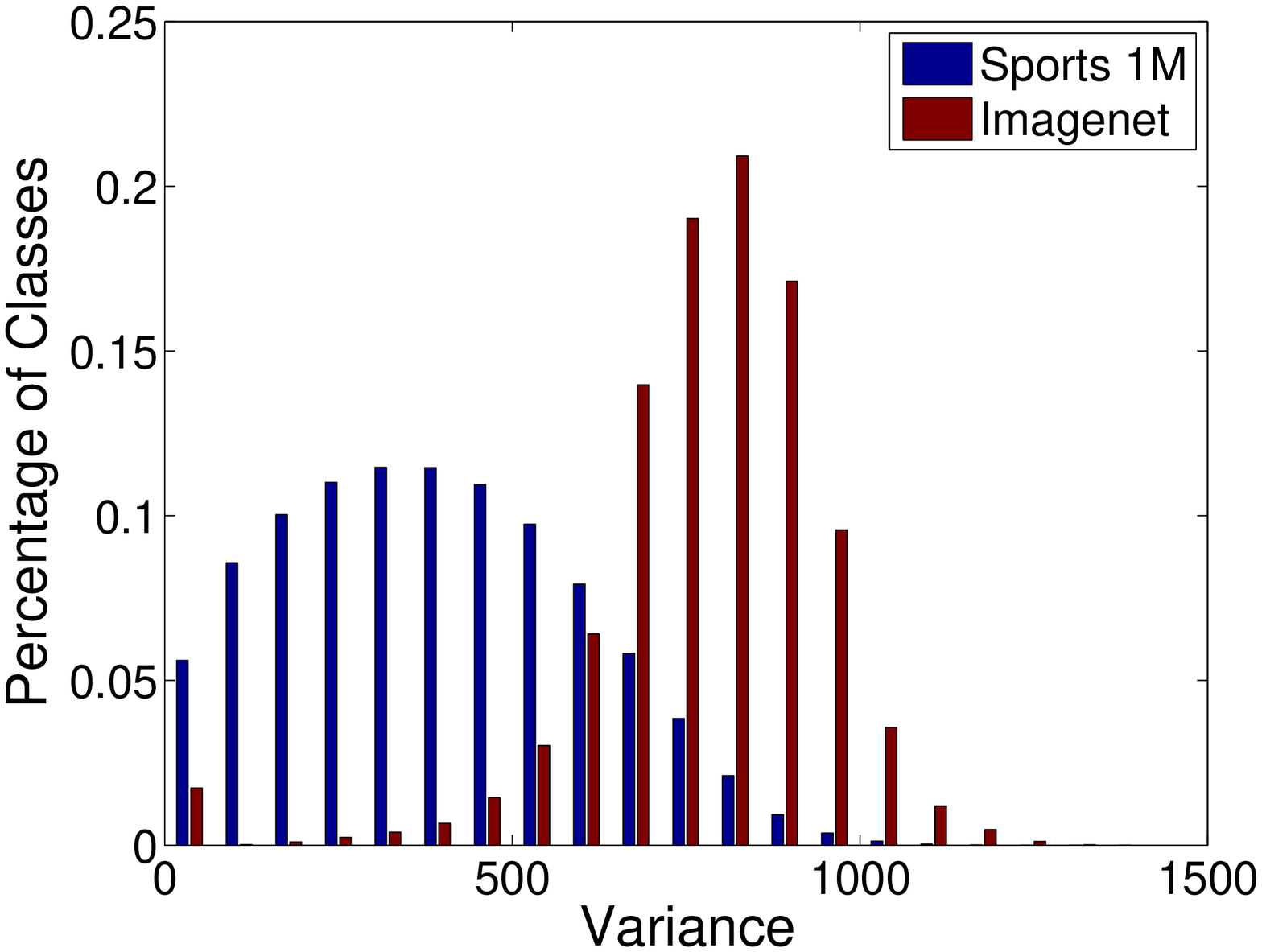}
\caption{\small The in-class variance of the examples in the Imagenet 21k and the Sports 1M datasets. The examples within a class are more spread out in the 
Imagenet 21k dataset than the Sports 1M dataset.}
\label{fig:variance}
\end{minipage}
\end{figure}

Figure~\ref{fig:sports-vs-steps} reports the accuracy on the evalution set against the number of steps trained for each model and
 Figure~\ref{fig:sports-vs-time} reports the accuracy against the actual time taken to complete these steps. We find that on both counts the WTA based 
model learns faster than both softmax and hierarchical softmax. 

The step time of the WTA model is about $4$ times lower than the softmax model but about $4$ higher than hierarchical softmax. This is because hierarchical 
softmax is much more efficient as it only computes $log(N)$ dot products compared to $K$ for WTA. However, even though hierarchical softmax processes 
significantly more number of examples the WTA models is able to achieve much higher accuracies.

In order to better understand the significant difference between WTA and the baselines on this task as opposed to the Imagenet 21K problem we computed 
the in-class variance of all the classes in the two datasets based on the $4000$-dim feature from the penultimate layer of the AlexNet model.
 Figure~\ref{fig:variance} reports a histogram of the in-class variance of the examples belonging to a class on the two datasets. We find that in the Imagenet
task the examples within a class are much more spread out than the Sports 1M task which is expected given that frames within a video would have similar context 
and more correlation. This could explain the relative efficiency of the top $K$ gradient updates used by the WTA model on the Sports 1M task.

\section{Conclusions}
\label{sec:conclusions}
We proposed a locality sensitive hashing approach for approximating the computation of $x^TW$ in the classification layer of deep network models
which enables us to scale up the training and inference of these models to millions of classes. Empirical evaluations of the proposed model on 
various large-scale datasets shows that the proposed approach provides significant speed-ups over baseline softmax models and can train such
large-scale models at a faster rate than alternatives such as hierarchical softmax. Our approach is advantageous whenever the number of classes considered 
is large or where batching is not possible.

In the future we would like to extend this technique to intermediate
layers also as the proposed method explicitly places sparsity constraints which is desirable in hierarchical learning. Given the scaling properties of hashing,
 our approach could, for instance, be used to increase the number of filters used in the convolutional layers from hundreds to tens of thousands with a few
 hundred being active at any time.

{\small
\bibliographystyle{iclr2015}
\bibliography{egbib}
}

\end{document}